\documentclass[IROS]{RLIpaper}

\usepackage{geometric_algebra}

\begin{document}

\title{Extending the Cooperative Dual-Task Space\\ in Conformal Geometric Algebra}
\author{\IEEEauthorblockN{Tobias Löw\IEEEauthorrefmark{1}\IEEEauthorrefmark{2} and Sylvain Calinon\IEEEauthorrefmark{1}\IEEEauthorrefmark{2}}
\IEEEauthorblockA{\IEEEauthorrefmark{1}Idiap Research Institute, Martigny, Switzerland}
\IEEEauthorblockA{\IEEEauthorrefmark{2}EPFL, Lausanne, Switzerland}
}

\maketitle

\begin{abstract}
    In this work, we are presenting an extension of the cooperative dual-task space (CDTS) in conformal geometric algebra. The CDTS was first defined using dual quaternion algebra and is a well established framework for the simplified definition of tasks using two manipulators. By integrating conformal geometric algebra, we aim to further enhance the geometric expressiveness and thus simplify the modeling of various tasks. We show this formulation by first presenting the CDTS and then its extension that is based around a cooperative pointpair. This extension keeps all the benefits of the original formulation that is based on dual quaternions, but adds more tools for geometric modeling of the dual-arm tasks. We also present how this CGA-CDTS can be seamlessly integrated with an optimal control framework in geometric algebra that was derived in previous work. In the experiments, we demonstrate how to model different objectives and constraints using the CGA-CDTS. Using a setup of two Franka Emika robots we then show the effectiveness of our approach using model predictive control in real world experiments.
\end{abstract}



\section{INTRODUCTION}
\label{sec:introduction}
    \addgrant{This work was supported by the State Secretariat for Education, Research and Innovation in Switzerland for participation in the European Commission’s Horizon Europe Program through the INTELLIMAN project (https://intelliman-project.eu/, HORIZON-CL4-Digital-Emerging Grant 101070136) and the SESTOSENSO project (http://sestosenso.eu/, HORIZON-CL4-Digital-Emerging Grant 101070310).}

    With the increasing desire to deploy robots in human environemnts, the need for robots to have human-like manipulation capabilities arises. One inherent ability that humans have is to manipulate objects using both their hands and arms, which is needed for example when objects that are either too large or too heavy need to be manipulated. 
    In order to match these capabilities and to be able to mimic them, robotic systems also need to be able to cooperatively control two arms in order to perform tasks in human environments.

    Apart from dual-arm systems being more human-like in terms of form factor, they also have some technical advantages. One can have the stiffness and strength of parallel manipulators combined with the flexibility and dexterity of serial manipulators \cite{leeSelfreconfigurableDualarmSystem1991}. Furthermore, since they increase the redundancy in the task-space due to their high number of degrees of freedom, they are better suited for intricate tasks that require a high manipulability such as screw assembly \cite{lizaahmadshauriAssemblyManipulationSmall2011} and dishwashing \cite{ogrenMultiObjectiveControl2012}. Other applications of bimanual systems are manipulating articulated objects \cite{almeidaCooperativeManipulationIdentification2018} or cables \cite{zhuDualarmRoboticManipulation2018}. More advantages and examples are listed in \cite{smithDualArmManipulation2012}. 

    Since many problems in robotics boil down to optimization problems that can be solved efficiently with various state-of-the-art solvers, it is of great interest to facilitate the modeling and increase the expressiveness of the formulations. Choosing the correct representation can make a huge difference in terms of how much prior knowledge we can embed into the formulation of those optimization problems. These are in robotics often very geometric, hence it is very beneficial to choose representations that intuitively allow to incoporate the geometry of the problem. For the case of dual-arm manipulation, the cooperative dual-task space (CDTS) \cite{adornoDualPositionControl2010} was proposed. This approach uses dual quaternion algebra (DQA), which not only unifies the treatment of position and orientation, it also allows the representation of various geometric primitives \cite{adornoRobotKinematicModeling}. These primitives can then be used to simplify the modeling of the tasks.

        \begin{figure}[!t]
            \centering
            \includegraphics[width=0.9\linewidth]{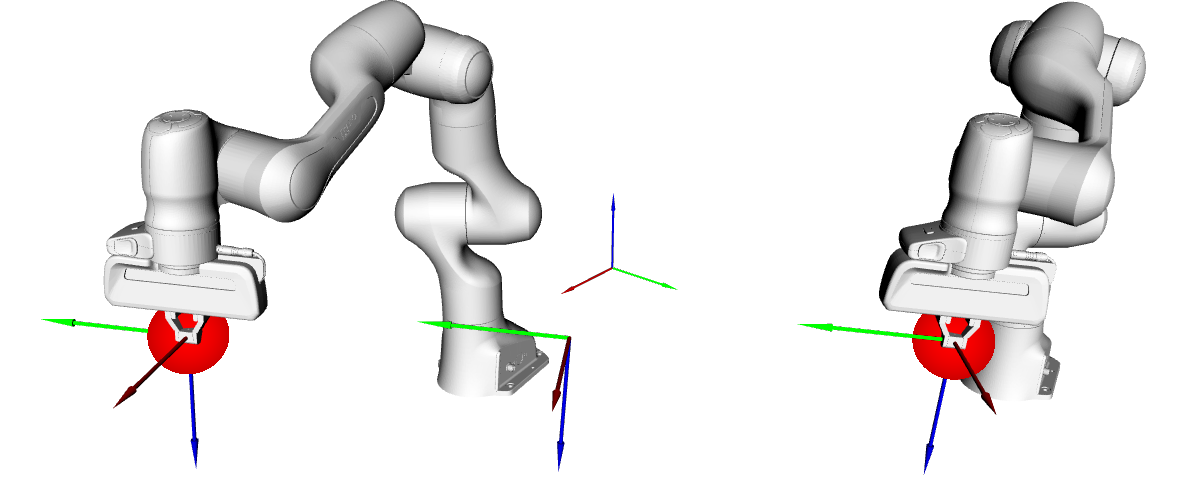}
            \caption{Cooperative dual-task space using conformal geometric algebra. The figure shows the two manipulators as well as their individual, relative and absolute motors. Additionally, it shows the cooperative pointpair.}
            \label{fig:cooperative_dual_task_space_}
        \end{figure}

    Dual quaternions have a strong connection to geometric algebra, especially the variants known as projective (PGA) and conformal (CGA) geometric algebra \cite{gunnGeometricAlgebrasEuclidean2017}, since dual quaternions are isomorphically embedded in their sub-algebras \cite{gunnProjectiveGeometricAlgebra2020}. The geometric algebras, however, are richer algebras that offer more geometric primitives and, more importantly, they offer the geometric construction of primitives based on operations such as intersections \cite{gunnProjectiveGeometricAlgebra2020}. This leads to new possibilities when formulating objectives and constraints based on the geometric primitives in the CGA-CDTS compared to the DQ-CDTS, which we will show in the experiments.

    In this article, we formulate the CDTS in conformal geometric algebra and show how this formulation naturally extends the DQ-CDTS. The resulting CGA-CDTS retains the same properties of a compact representation of two arm system as the DQ-CDTS, while adding a useful geometric primitive, the cooperative pointpair, that represents both end-effector positions simultaneously. Furthermore, we demonstrate how the CGA-CDTS can be used in the optimal control formulation with geometric primitives for manipulators that we presented in \cite{lowGeometricAlgebraOptimal2023}. Hence, this article aims to explain the basic mathematical formulations of various optimization problems using the CGA-CDTS.


\section{BACKGROUND}
\label{sec:background}
    
    
    \subsection{Cooperative Dual-Task Space}
    \label{sub:cooperative_dual_task_space}
        We will introduce the cooperative dual-task space here conceptually, and not mathematically, since its original definition uses dual quaternion algebra. In contrast to that, we will be defining and extending it using conformal geometric algebra. So, for the sake of conciseness, we are keeping this introduction on a high level.

        The cooperative dual-task space was proposed as a compact and singularity-free representation of a two-arm system \cite{adornoDualPositionControl2010}. It is defined using two poses that depend on the end-effector poses of the two manipulators, one is the relative and the other one is the absolute pose. All poses are represented using unit dual quaternions, which have advantages compared to other representations, such as coupled position and orientation, singularity-free representation and efficient computation \cite{sariyildizComparisonStudyThree2011}. 

        Based on the compact representation of the CDTS, various control strategies have been proposed. In \cite{fonsecaCoupledTaskSpaceAdmittance2020}, a coupled task-space admittance controller was presented, that allowed for a geometrically consistent stiffness term. A reactive control strategy was developed in \cite{lahaReactiveCooperativeManipulation2021} that leveraged geometric primitves for task relaxations and priorities. Task priorities for control in the CDTS were also proposed in \cite{defariasDesignSingularityrobustTaskpriority2017}. In order to exploit human demonstration that allow the teaching of cooperative motions, motion primitives for bimanual systems were presented based on the CDTS \cite{vorndammeIntegratedBiManualMotion2022a}.

        Note that the results of the research that is based on the DQ-CDTS can also be used with the CGA-CDTS, albeit with mathematical changes due to the different algebra. Furthermore, the mentioned advantages of DQA also apply to CGA, since dual quaternions and the corresponding subalgebra in CGA, i.e. the motors, are isomorphic \cite{bayro-corrochanoMotorAlgebraApproach2000}.

    
    \subsection{Geometric Algebra}
    \label{sub:geometric_algebra}
        Geometric algebra is a single algebra for geometric reasoning, alleviating the need of utilizing multiple algebras to express geometric relations. In this article, we are using the variant known as conformal geometric algebra (CGA). We use the following notation: $x$ to denote scalars, $\bm{x}$ for vectors, $\bm{X}$ for matrices, $X$ for multivectors and $\bm{\mathcal{X}}$ for matrices of multivectors.

        A general element for computation in geometric algebra is called a multivector. There are three main products that can be used with multivectors: the geometric product $XY$, the inner product $X \inner Y$ and the outer product $X\outer Y$. The trivial vector case shows that the geometric product combines the inner $\cdot$ and the outer $\outer$ product
        \begin{equation}\label{eq:geometric_product}
            \geometricproduct.
        \end{equation}
        
        The outer product is a spanning operation that allows the creation of geometric primitives from points $P_i$. For example, two points $P_1\outer P_2$ yield a point pair, three points $P_1\outer P_2 \outer P_3$ a circle and four points $P_1\outer P_2 \outer P_3\outer P_4$ a sphere. There are more primitives, that we will introduce when we use them in the experiments.

        Rigid body transformations are represented by motors $M$. They can be applied to multivectors by a sandwiching operation, similar to how quaternions rotate vectors        
        \begin{equation}\label{eq:motor_transforming_multivector}
            Y = MX\reverse{M},
        \end{equation}
        where $\reverse{M}$ stands for the reverse of a motor. Motors are exponential mappings of so-called bivector (i.e. the subspaces spanned by the outer product of two vectors), the inverse operation is the logarithmic map
        \begin{equation}\label{eq:motor_exp_log_map}
            M = \exp(B)
            \hspace{3mm}
            \iff
            \hspace{3mm}
            B = \log(M).
        \end{equation}        
        The motor $M (\posjoint)$  corresponding to the forward kinematics of a kinematic chain can be computed as the product of the individual joint motors
        \begin{equation}\label{eq:cgaforwardkinematics}
            M (\posjoint) = \prod_{i=1}^{N} M_i(q_i).
        \end{equation}

        The analytic Jacobian can then be found as the derivative of the forward kinematics motor defined in Equation \eqref{eq:cgaforwardkinematics}, i.e.
        \begin{equation}\label{eq:ga_analytic_jacobian}
            \gaajacobian = \frac{\partial M(\posjoint)}{\partial \posjoint} = \mat{\frac{\partial M(\posjoint)}{\partial q_1} \ldots \frac{\partial M(\posjoint)}{\partial q_N}}.
        \end{equation}

        At this point, we only introduced the most important concepts for understanding the proposed method. For a complete introduction we refer interested readers to \cite{perwassGeometricAlgebraApplications2009} and \cite{bayro-corrochanoGeometricAlgebraApplications2020}.


\section{METHOD}
\label{sec:method}
    The CDTS in conformal geometric algebra is an extension of the CDTS in dual quaternion algebra. We first present the basic reformulation of the CDTS in CGA and then its extension by using the additional geometric primitives. Lastly, we present how the CGA-CDTS can be used within an optimal control framework using geometric algebra that we previously proposed for manipulation tasks.
    
    
    \subsection{Conformal Geometric Algebra Cooperative Dual-Task Space}
    \label{sub:conformal_geometric_algebra_cooperative_dual_task_space}
        The geometric algebra equivalent of the relative and absolute dual quaternions of the DQ-CDTS are defined using motors. Given the joint configurations of the two manipulators, $\posjoint_1$ and $\posjoint_2$, respectively, we can easily find their end-effector motors $M_1(\posjoint_1)$ and $M_2(\posjoint_2)$ using the forward kinematics in CGA. From this it is straightforward to formulate the relative motor as
        \begin{equation}\label{eq:cdts_relative_motor}
            M_r(\posjoint_1,\posjoint_2) = \reverse{M}_2(\posjoint_2) M_1(\posjoint_1),
        \end{equation}
        while its Jacobian, i.e. the relative analytic Jacobian, can be found as
        \begin{equation}\label{eq:relative_analytic_jacobian}
            \gamatrix{J}^A_r(\posjoint_1,\posjoint_2) = 
            \mat{
                \reverse{M}_2(\posjoint_2) \gamatrix{J}^A_1(\posjoint_1)
                &
                \gamatrix{\reverse{J}}^A_2(\posjoint_2) M_1(\posjoint_1) 
            }.
        \end{equation}
        
        Similarly, the absolute motor can be found as
        \begin{equation}\label{eq:cdts_absolute_motor}
            \begin{split}
                M_a(\posjoint_1,\posjoint_2) 
                & = M_2(\posjoint_2) M_{r/2}(\posjoint_1,\posjoint_2)
                \\
                & = M_2(\posjoint_2)\exp\left(\frac{1}{2}\log\Big(M_r(\posjoint_1,\posjoint_2)\Big)\right),
            \end{split}
        \end{equation}
        with its corresponding absolute analytic Jacobian
        \begin{equation}\label{eq:absolute_analytic_jacobian}
            \begin{split}
                \gamatrix{J}^A_a(\posjoint_1,\posjoint_2) = & M_2(\posjoint_2) \gamatrix{J}^A_{M_{r/2}}(\posjoint_1,\posjoint_2) 
                \\        
                & + \mat{
                    \bm{0} & \gamatrix{J}^A_2(\posjoint_2) M_{r/2}(\posjoint_1,\posjoint_2)
                },
            \end{split}
        \end{equation}
        where $\gamatrix{J}^A_{M_{r/2}}(\posjoint_1,\posjoint_2)$ is the analytic Jacobian of the motor $M_{r/2}(\posjoint_1,\posjoint_2)$. It can be found as 

        \begin{equation}\label{eq:jacobian}
            \gamatrix{J}^A_{M_{r/2}}(\posjoint_1,\posjoint_2) = \bm{J}_{\mathbb{B}\to\mathcal{M}}(B_{r/2})\bm{J}_{\mathcal{M}\to\mathbb{B}}(M_{r})\bm{J}^A_r(\posjoint_1,\posjoint_2).
        \end{equation}
        Here, $B_{r/2}$ is the logarithm of the motor $M_{r/2}$. The matrices $\bm{J}_{\mathbb{B}\to\mathcal{M}}$ and $\bm{J}_{\mathcal{M}\to\mathbb{B}}$ are the Jacobians of the exponential and logarithmic mapping respectively. We already showed the derivation of the Jacobian of the logarithmic mapping in the appendix of \cite{lowGeometricAlgebraOptimal2023}. The Jacobian of the exponential mapping can be found in Appendix A.

        Both the relative and the absolute analytic Jacobians are $1\times 2N$ multivector matrices that contain motors as their elements. Hence, when expanding it to normal matrix algebra they become $8\times 2N$ matrices.

        Since motors in CGA can be used to transform any geometric primitive that is part of the algebra in a uniform way, it is easy to find cooperative geometric primitives. Their definition can be trivially found using Equation \eqref{eq:motor_transforming_multivector}, where $M$ is either the relative $M_r(\posjoint_1,\posjoint_2)$ or absolute $M_a(\posjoint_1,\posjoint_2)$ motor and $X$ can be any geometric primitive. The corresponding Jacobians are then found using the respective Jacobians $\gamatrix{J}^A_r(\posjoint_1,\posjoint_2)$ and $\gamatrix{J}^A_a(\posjoint_1,\posjoint_2)$. 

    
    \subsection{Cooperative Pointpair}
    \label{sub:cooperative_pointpair}
        In extension to the CDTS that was defined using dual quaternion algebra, the CDTS presented here using CGA also allows a geometric primitive that corresponds to both end-effector positions simultaneously. This cooperative pointpair is defined as the outer product of the two end-effector points, i.e.
        \begin{equation}\label{eq:cooperative pointpair}
            P_{cdts} = M_1(\posjoint_1)\gae{0}\reverse{M}_1(\posjoint_1) \outer M_2(\posjoint_2)\gae{0}\reverse{M}_2(\posjoint_2).
        \end{equation}
        The Jacobian of the cooperative pointpair can be found as
        \begin{equation}\label{eq:jacobian_cooperative_pointpair}
            \gamatrix{J}_{P_{cdts}} = \mat{
                \gamatrix{J}_{P_{cdts},1}
                &
                \gamatrix{J}_{P_{cdts},2}
                
            },
        \end{equation}
        where
        \begin{equation}\label{eq:jacobian_cooperative_pointpair_1}
            \begin{split}
                \gamatrix{J}_{P_{cdts},1} =    
                & \gamatrix{J}^A_1(\posjoint_1)\gae{0}\reverse{M}_1(\posjoint_1) \outer M_2(\posjoint_2)\gae{0}\reverse{M}_2(\posjoint_2)
                \\ 
                & + M_1(\posjoint_1)\gae{0}\gamatrix{\reverse{J}}^A_1(\posjoint_1) \outer M_2(\posjoint_2)\gae{0}\reverse{M}_2(\posjoint_2),
            \end{split}
        \end{equation}
        and
        \begin{equation}\label{eq:jacobian_cooperative_pointpair_2}
            \begin{split}
                \gamatrix{J}_{P_{cdts},2} =    
                & M_1(\posjoint_1)\gae{0}\reverse{M}_1(\posjoint_1) \outer \gamatrix{J}^A_2(\posjoint_2)\gae{0}\reverse{M}_2(\posjoint_2) 
                \\
                & + M_1(\posjoint_1)\gae{0}\reverse{M}_1(\posjoint_1) \outer M_2(\posjoint_2)\gae{0}\gamatrix{\reverse{J}}^A_2(\posjoint_2).
            \end{split}
        \end{equation}

\begin{figure}[!ht]
            \centering
            \includegraphics[width=0.9\linewidth]{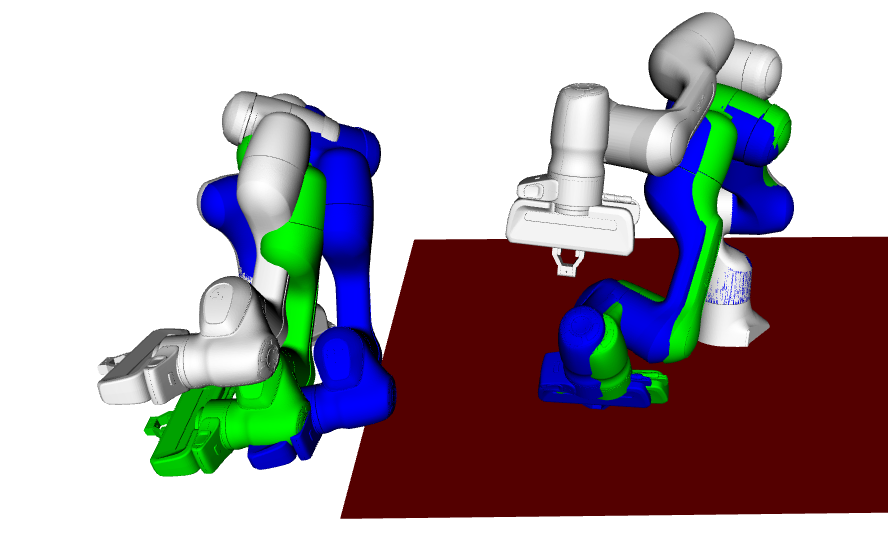}
            \caption{Dual-Arm manipulator reaching a plane. The target plane is shown in red. The white Franka Emika robots show the initial configurations, the green ones are the result of individually reaching the plane, and the blue ones cooperatively.}
            \label{fig:dual_arm_manipulator_reaching_a_plane_}
        \end{figure}

        Note that the cooperative pointpair is a direct representation of both points and is not the same as stacking the two points. Therefore the Jacobian matrix is also different, which will lead to different solutions of optimization problems. An example of this is shown in Figure \ref{fig:dual_arm_manipulator_reaching_a_plane_}, where two Franka Emika robots are tasked to reach a plane, once individually (i.e. by stacking their end-effector points) and once cooperatively (i.e. by using the cooperative pointpair that is presented here). It can be seen that the corresponding solution configurations are not the same, which shows that the cooperative pointpair representation lets the two robots influence each other. 

        Evidently, this cooperative pointpair Jacobian has a singularity in the case when both end-effector positions are equal. This will cause the outer product, by definition, to be zero. In practice, this can be avoided by posing a constraint on the distance between end-effectors, which usually is required anyways.

    
    \subsection{Using the CGA-CDTS for Optimal Control}
    \label{sub:using_the_cga_cdts_for_optimal_control}
        Integrating the CGA-CDTS into the optimal control framework that we presented in \cite{lowGeometricAlgebraOptimal2023} can be achieved by replacing the single end-effector motor with the relative and absolute motors, respectively. The objective of reaching a target motor is then formulated as 

        \begin{equation}\label{eq:reaching_target_motor}
            E_M(\posjoint_1,\posjoint_2) = \log \left( \reverse{M}_{target} M(\posjoint_1,\posjoint_2) \right),
        \end{equation}
        where $M(\posjoint_1,\posjoint_2)$ can be either $M_r(\posjoint_1,\posjoint_2)$ or $M_a(\posjoint_1,\posjoint_2)$.
        
        Alternatively, we can define a residual multivector for reaching a geometric primitive $X_d$ as
        \begin{equation}
            E_{X_d}(\posjoint_1,\posjoint_2) = X_{d} \outer M(\posjoint_1,\posjoint_2) X \reverse{M}(\posjoint_1,\posjoint_2),
        \end{equation}
        again using either the relative or absolute motor. Deriving the respective Jacobians is straightforward using the definitions of the relative and absolute analytic Jacobians in Equations \eqref{eq:relative_analytic_jacobian} and \eqref{eq:absolute_analytic_jacobian}. With this, these residual multivectors of the CGA-CDTS can then directly be used to define objectives or constraints for optimal control problems, which would mean for example that a relative or absolute geometric primitive should be reached. 

        The cooperative pointpair can also be used in order to define tasks as optimization problems. The first way to do so is using the outer product in a similar way to the above relative and absolute residual multivectors, i.e. 
        \begin{equation}\label{eq:cooperative_pointpair_objective}
            E_{cdts}(\posjoint_1,\posjoint_2) = P_{target} \outer P_{cdts}(\posjoint_1,\posjoint_2),
        \end{equation}
        with the Jacobian 
        \begin{equation}\label{eq:cooperative_pointpair_objective_jacobian}
            \gamatrix{J}_{E_{cdts}}(\posjoint_1,\posjoint_2) = P_{target} \outer \gamatrix{J}_{P_{cdts}}(\posjoint_1,\posjoint_2),
        \end{equation}
        This objective means that the two manipulators should cooperatively reach a single point. The implications and results of this objective are further detailed in the experiment section. 
        
        Another common use-case are containment relationships for the cooperative pointpair with respect to other geometric primitives. These can then be used to define tasks where the dual-arm system should cooperatively reach a target. The mathematical formulation is using a product called the commutator product $\times$, i.e. 
        \begin{equation}\label{eq:commutator_product}
            \begin{split}
                E_{cdts}(\posjoint_1,\posjoint_2) 
                & = X_d \times P_{cdts}(\posjoint_1,\posjoint_2) 
                \\
                & = \frac{1}{2}(X_dP_{cdts}(\posjoint_1,\posjoint_2) - P_{cdts}(\posjoint_1,\posjoint_2)X_d).
            \end{split}
        \end{equation}
        The Jacobian of this containment relationship can be found as
        \begin{equation}\label{eq:jacobian_containment}
            \gamatrix{J}_{E_{cdts}}(\posjoint_1,\posjoint_2) = X_d \times \gamatrix{J}_{E_{cdts}}(\posjoint_1,\posjoint_2).
        \end{equation}
        The interpretation of the residual multivector $E_{cdts}(\posjoint_1,\posjoint_2)$ is that it should be reduced to zero if the cooperative pointpair is contained within the desired geometric primitive $X_d$, e.g. if both end-effector points lie on a circle.

        The distance between the two end-effector can be constrained using the inner product between the two end-effector points, i.e.
        \begin{equation}\label{eq:distance_constraint}
            E_d(\posjoint_1,\posjoint_2) = -2M_1(\posjoint_1)\gae{0}\reverse{M}_1(\posjoint_1) \inner M_2(\posjoint_2)\gae{0}\reverse{M}_2(\posjoint_2)-d^2,
        \end{equation}
        where $d$ is the desired distance.        


\section{EXPERIMENTS}
\label{sec:experiments}
    In this section, we are presenting various cooperative tasks that are defined in the CGA-CDTS. For each task we are providing the mathematical definition of the optimization problem. We are then solving those optimization problems using standard solvers such as Gauss-Newton. For simplicity, the problems in simulation are formulated essentially as inverse kinematics problems, the same objectives and constraints can, however, be used in optimal control problems that are for example then used for model predictive control. We are demonstrating this in the real world experiments, where the problems are then solved by using a variant of the iterative linear quadratic regulator (iLQR) \cite{tassaSynthesisStabilizationComplex2012}. Both the simulation and the real-world experiments use the same setup of two table-top mounted Franka Emika robots. Additional material for the experiments as well as the videos of the real world experiments can be found on the accompanying website \footnote{\url{https://geometric-algebra.tobiloew.ch/cdts/}}. All geometric algebra computations are done using our open-source library \textit{gafro}\footnote{\url{https://github.com/idiap/gafro}}.

    
    \subsection{Cooperatively Reaching a Point}
    \label{sub:cooperatively_reaching_a_point}
        As mentioned before, the cooperative pointpair models both end-effector positions simultaneously. Hence, when formulating a simple optimization problem such as reaching a single point, the system automatically chooses which manipulator should perform the task. Since the information of both manipulators is encoded in one geometric primitive, it alleviates the need to construct the problem using conditional formulations. The problem can be expressed compactly using the outer product, i.e. 
        \begin{equation}\label{eq:cooperatively_reaching_point}
            \begin{split}
                \posjoint^* = \min_{\posjoint}& \Big\| P_{target} \outer P_{cdts} (\posjoint_1,\posjoint_2) \Big\|_2^2.
            \end{split}
        \end{equation}

\begin{figure}[!ht]
            \centering
            \begin{subfigure}[t]{\linewidth}
                \includegraphics[width=0.9\linewidth]{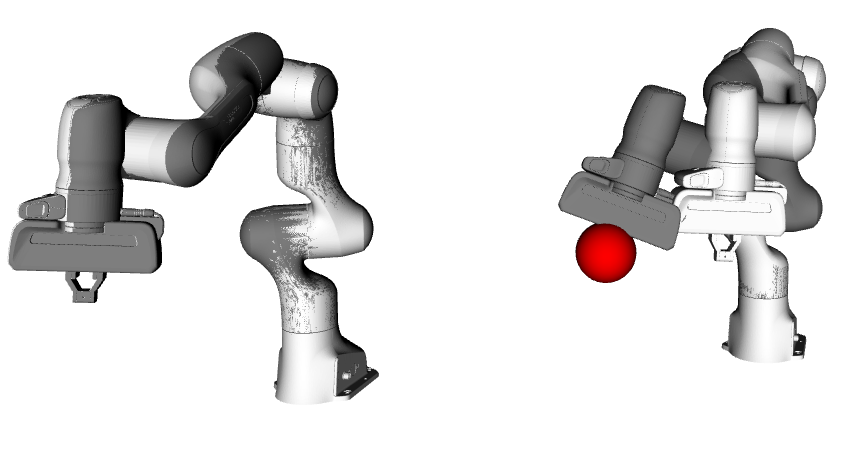}
                \caption{Left arm reaching for the point.}
            \end{subfigure}
            \begin{subfigure}[t]{\linewidth}
                \includegraphics[width=0.9\linewidth]{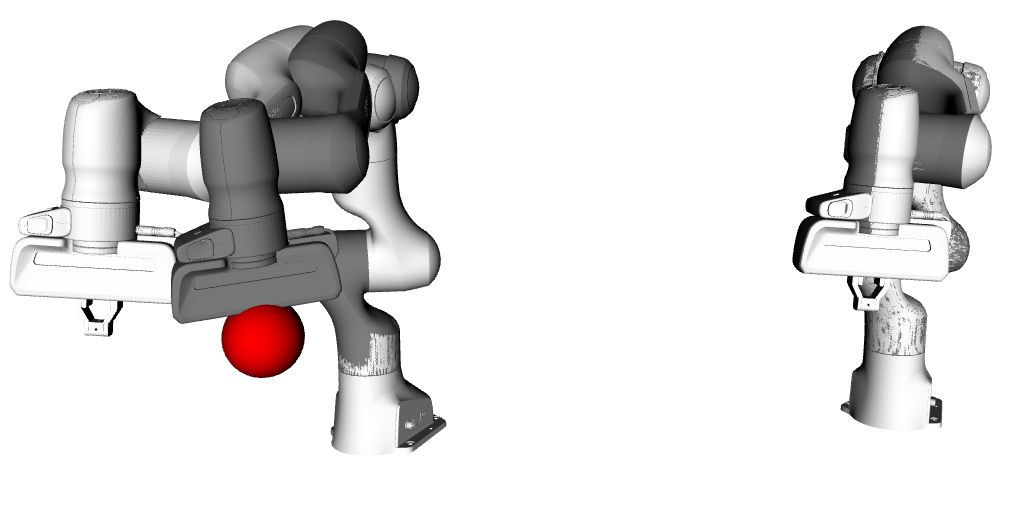}
                \caption{Right arm reaching for the point.}
            \end{subfigure}
            \caption{Two Franka Emika robots reaching for a single point in the cooperative dual-task space. The initial configurations are shown in white and the final ones in gray. The target point is shown in red.}
            \label{fig:two_franka_emika_robots_reaching_a_point_in_the_cooperative_dual_task_space_}
        \end{figure}

        An example of this task is shown in Figure \ref{fig:two_franka_emika_robots_reaching_a_point_in_the_cooperative_dual_task_space_}. Notice how in both cases the manipulator that is closer to the point performs the reaching task while the other remains in its initial configuration.

        We also show this experiment in using the real-world setup and the accompanying video can be found on our website.

    
    \subsection{Cooperatively Reaching a Circle}
    \label{sub:cooperatively_reaching_a_circle}
        One of the geometric primitives that is available in CGA but not DQA is a circle. Hence, we can use the CGA-CDTS to define a task where a dual arm system should cooperatively reach a circle. An example application of reaching a circle would be holding a filled bucket with two manipulators. Mathematically, a circle is obtained by the outer product of three points. 

        The problem of cooperatively reaching a circle is formulated as a constrained optimization problem using the cooperative pointpair
        \begin{equation}\label{eq:cooperatively_reaching_circle}
            \begin{split}
                \posjoint^* = \min_{\posjoint}& \Big\| \log \left( \reverse{M}_r(\posjoint_0) M_r(\posjoint) \right) \Big\|_2^2
                \\
                & \text{s.t. } C \times P_{cdts} (\posjoint_1,\posjoint_2) = 0.
            \end{split}
        \end{equation}

\begin{figure}[!ht]
            \centering
            \includegraphics[width=0.9\linewidth]{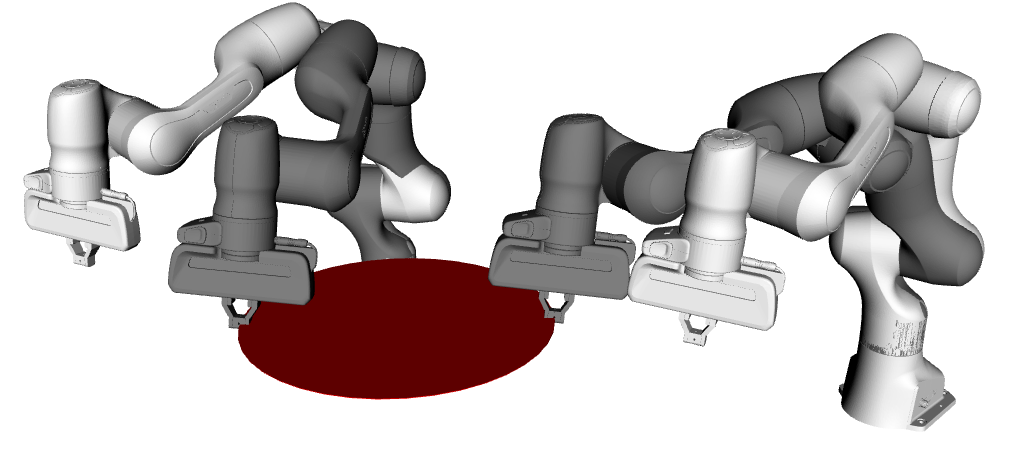}
            \caption{Two Franka Emika robots reaching a circle in the cooperative dual-task space while trying to maintain the same relative motor.}
            \label{fig:two_franka_emika_robots_reaching_a_circle_in_the_cooperative_dual_task_space_}
        \end{figure}

        This optimization problem formulates the task of reaching a circle, while trying to maintain the initial relative motor. The result of this problem can be seen in Figure \ref{fig:two_franka_emika_robots_reaching_a_circle_in_the_cooperative_dual_task_space_}.
        Formulating the task in this manner circumvents the mentioned singularity issue in the Jacobian of the cooperative pointpair, since the system will reach circle while keeping as close as possible to their initial relative poses.
        
    
    \subsection{Aligning Orientation Axis}
    \label{sub:aligning_orientation_axis}
        When two manipulators are cooperatively manipulating an object, often it is necessary to partially constrain their relative orientation, which is necessary in nearly all dual-arm grasping scenarios of rigid objects. One way to do so is enforcing collinear lines in the desired direction at the end-effector level. This is again achieved by using the commutator product, i.e. 

        \begin{equation}\label{eq:aligning_endeffector_lines}
            \begin{split}
                E_{L_1L_2} & (\posjoint_1,\posjoint_2)
                \\
                & = M_1(\posjoint_1) L_1 \reverse{M}_1(\posjoint_1) \times M_2(\posjoint_2) L_2 \reverse{M}_2(\posjoint_2).
            \end{split}
        \end{equation}

\begin{figure}[!ht]
            \centering
            \includegraphics[width=0.9\linewidth]{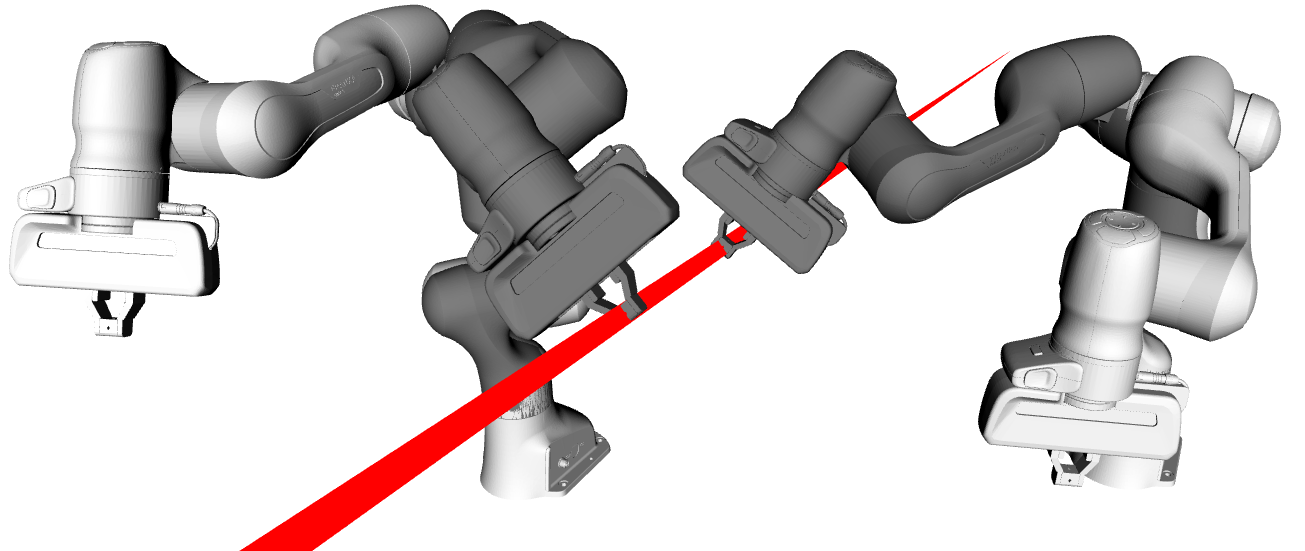}
            \caption{Aligning the x-Axis.}
            \label{fig:aligning_the_}
        \end{figure}

        In Figure \ref{fig:aligning_the_}, we show the  results of minimizing $E_{L_1L_2}(\posjoint_1,\posjoint_2)$, where $L_1=L_2=\gae{0}\outer(\gae{0}+\gae{1}+\frac{1}{2}\gae{\infty})\outer \gae{\infty}$, which corresponds to aligning two lines that pass through the x-axes of the frames at the end-effectors of the two manipulators.

        There are, of course, other ways to achieve this behavior, e.g. by constraining the orientation part of the relative motor. We chose to demonstrate this method of aligning orientation, however, because it provides a lot of flexibility, since the two lines $L_1$ and $L_2$ can be chosen arbitrarily.

        A similar instance of this problem can be defined using the absolute motor of the CGA-CDTS. By using the absolute translator $T_a(\posjoint_1,\posjoint_2)$ we can move a desired line to the absolute position and require it to be identical to the line moved by the absolute motor. This defines a desired orientation with respect to the arbitrary axis that is defined by the line. The formulation hence is
        \begin{equation}\label{eq:constraining_absolute_axis}
            \begin{split}
                E_{T_L} & (\posjoint_1,\posjoint_2)
                \\
                & = T_a(\posjoint_1,\posjoint_2)L\reverse{T}_a(\posjoint_1,\posjoint_2) \times M_a(\posjoint_1,\posjoint_2)L\reverse{M}_a(\posjoint_1,\posjoint_2).
            \end{split}
        \end{equation}

    
    \subsection{Balancing a Plate}
    \label{sub:reactively_balancing_plate_with_glass_of_water}
        In this real-world experiment we are combining several of the previous constraints in order to implement the task of balancing a plate. First, the robot lifts the plate to a height of 20cm above the table, then it will try to keep it in that position. This is formulated as constrained optimization problem, where the objective is to stay close to the initial configuration and the constraint is to keep the $z$-axis of absolute motor perpendicular to the $xy$-plane.
        
        We formulate this task as an optimal control problem
        \begin{equation}\label{eq:balancing_plane}
            \begin{split}
                \bm{u}^* = &\min_{\bm{u}} \Big\|E_N(\bm{x}_N)\Big\|_2^2 + \sum_{k=0}^{N-1} \Big\|E_k(\bm{x}_k)\Big\|_2^2 + \|\bm{u}_k\|_R^2
                \\
                & \text{s.t. } \bm{x}_{k+1} = f(\bm{x}_k,\bm{u}_k)
            \end{split}
        \end{equation}
        where $\bm{x} = \transpose{\mat{\transpose{\posjoint}_1, \transpose{\posjoint}_2, \transpose{\veljoint}_1, \transpose{\veljoint}_2}}$ and $\bm{u} = \transpose{\mat{\transpose{\accjoint}_1, \transpose{\accjoint}_2}}$. As the state dependent cost we choose the residual multivectors of Equations \eqref{eq:aligning_endeffector_lines} and \eqref{eq:constraining_absolute_axis}, where the line $L$ is chosen to be the $z$-axis, i.e. $L=\gae{0}\outer(\gae{0}+\gae{3}+\frac{1}{2}\gae{\infty})\outer \gae{\infty}$. This means that the two manipulators should simultaneously keep their relative grasping positions on the plate and to keep the plate horizontal.

        This formulation is given to a model predictive controller that uses second order system dynamics to compute desired accelerations. Using inverse dynamics we then compute torque commands for the control of the two manipulators. We chose a short horizon of 10 timesteps with $\Delta t = 0.01$, in order to achieve a very reactive behavior of the controller. The model predictive controller is then run at 100Hz and the inverse dynamics controller at 1000Hz.
        
        \begin{figure}[!ht]
            \centering
            \begin{subfigure}[t]{0.49\linewidth}
                    \includegraphics[width=\linewidth]{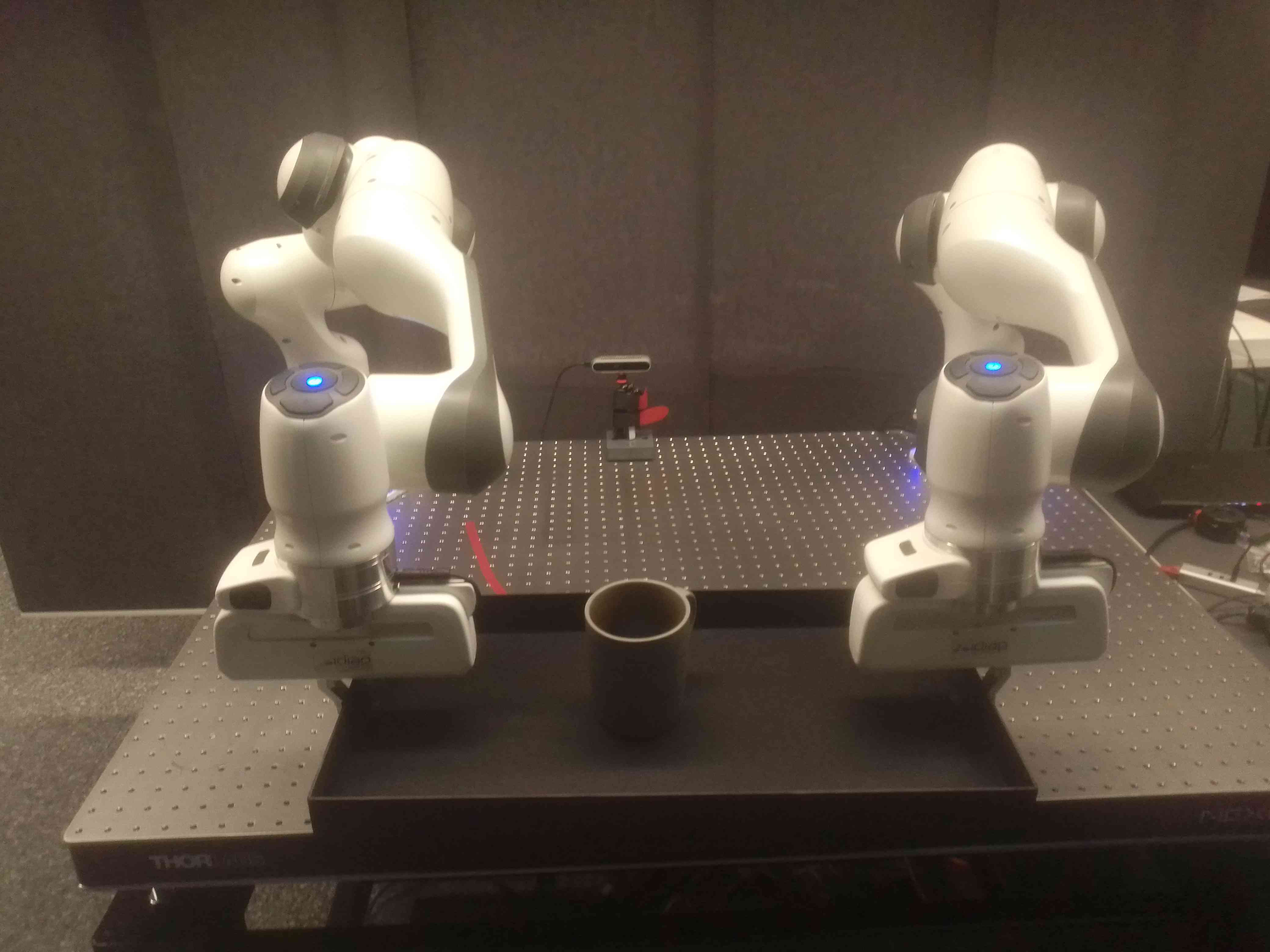}
            \end{subfigure}
            \hfill
            \begin{subfigure}[t]{0.49\linewidth}
                    \includegraphics[width=\linewidth]{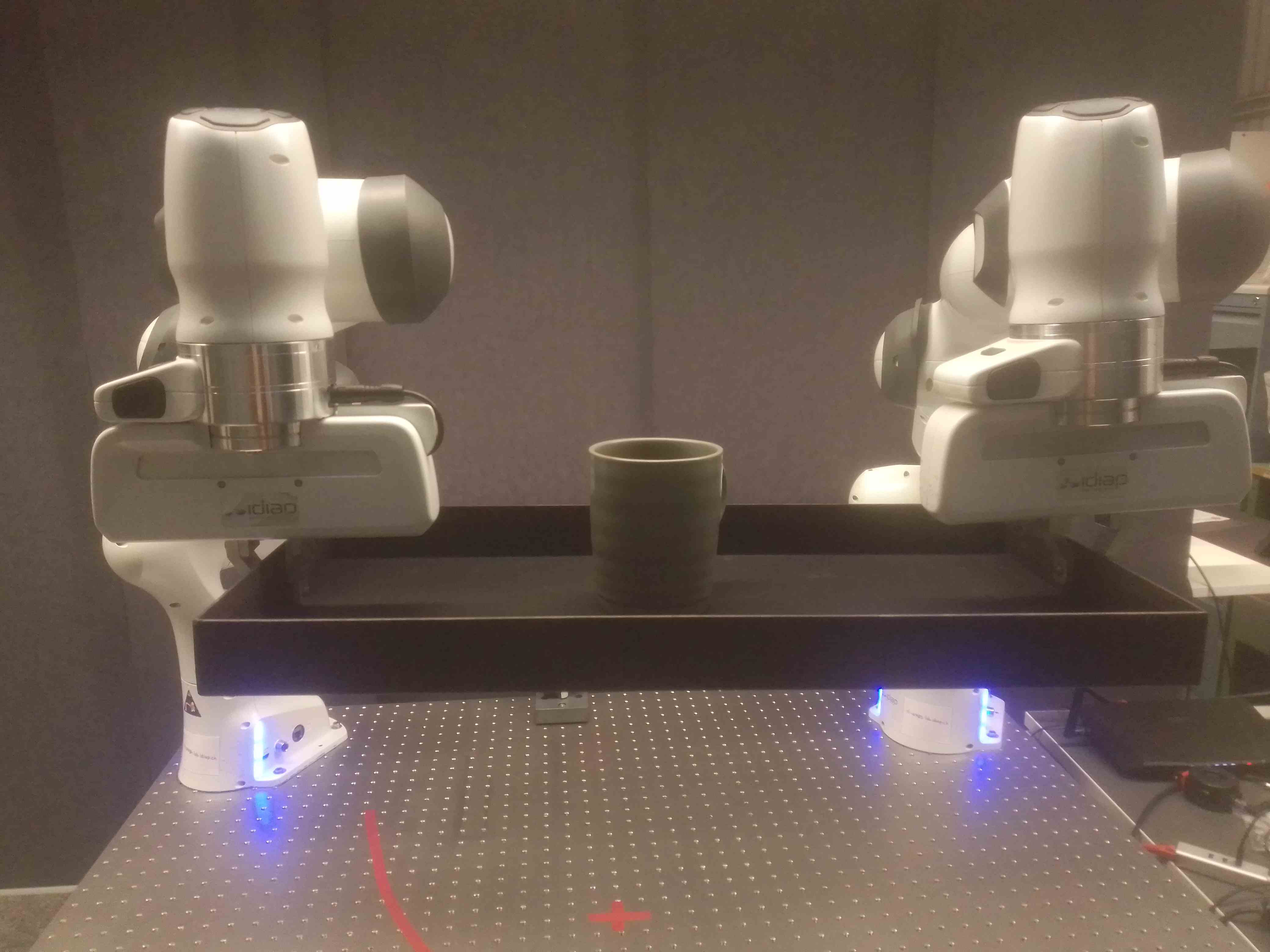}
            \end{subfigure}
            \\[0.02\linewidth]
            \begin{subfigure}[t]{0.49\linewidth}
                    \includegraphics[width=\linewidth]{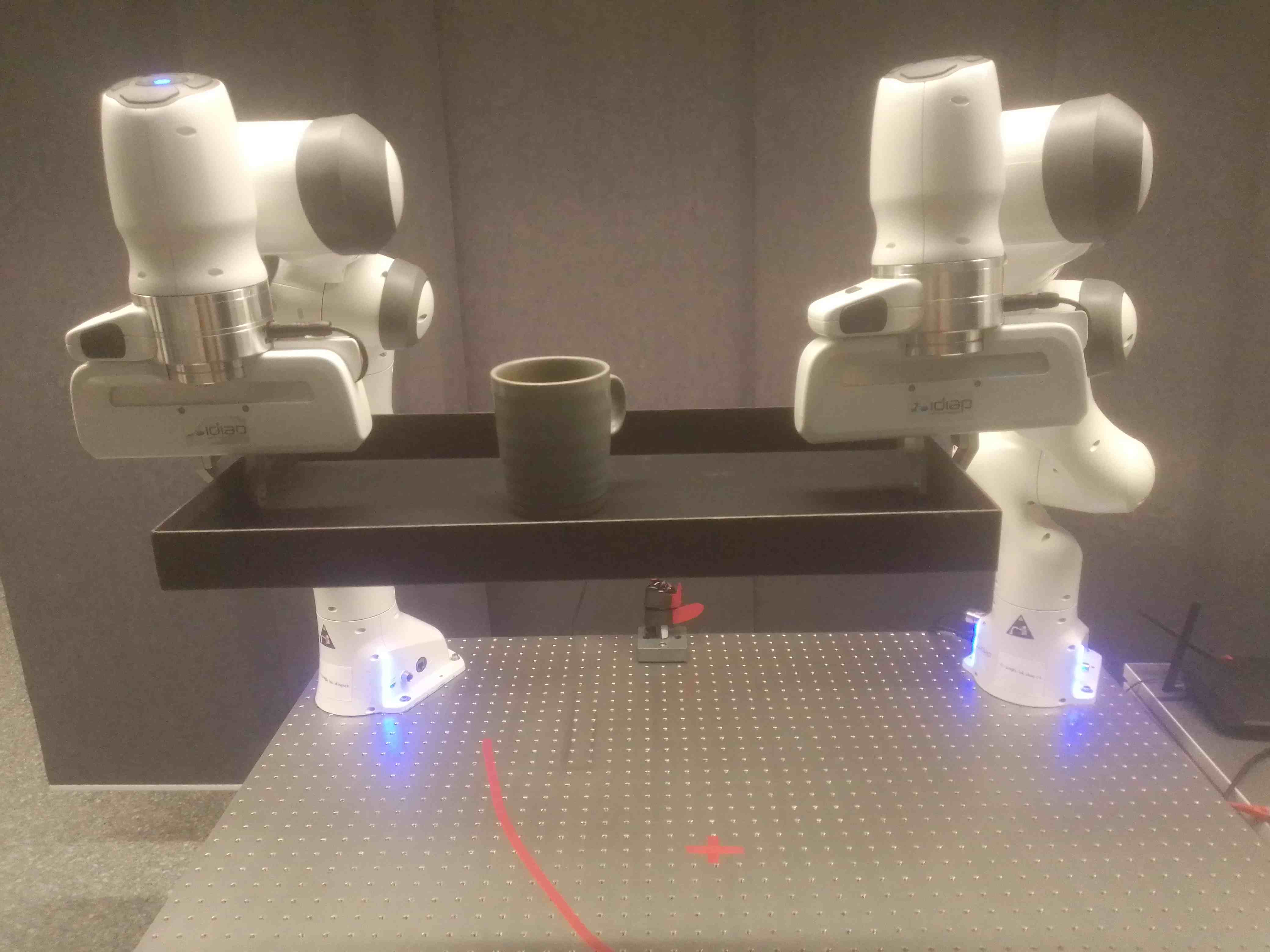}
            \end{subfigure}
            \hfill
            \begin{subfigure}[t]{0.49\linewidth}
                    \includegraphics[width=\linewidth]{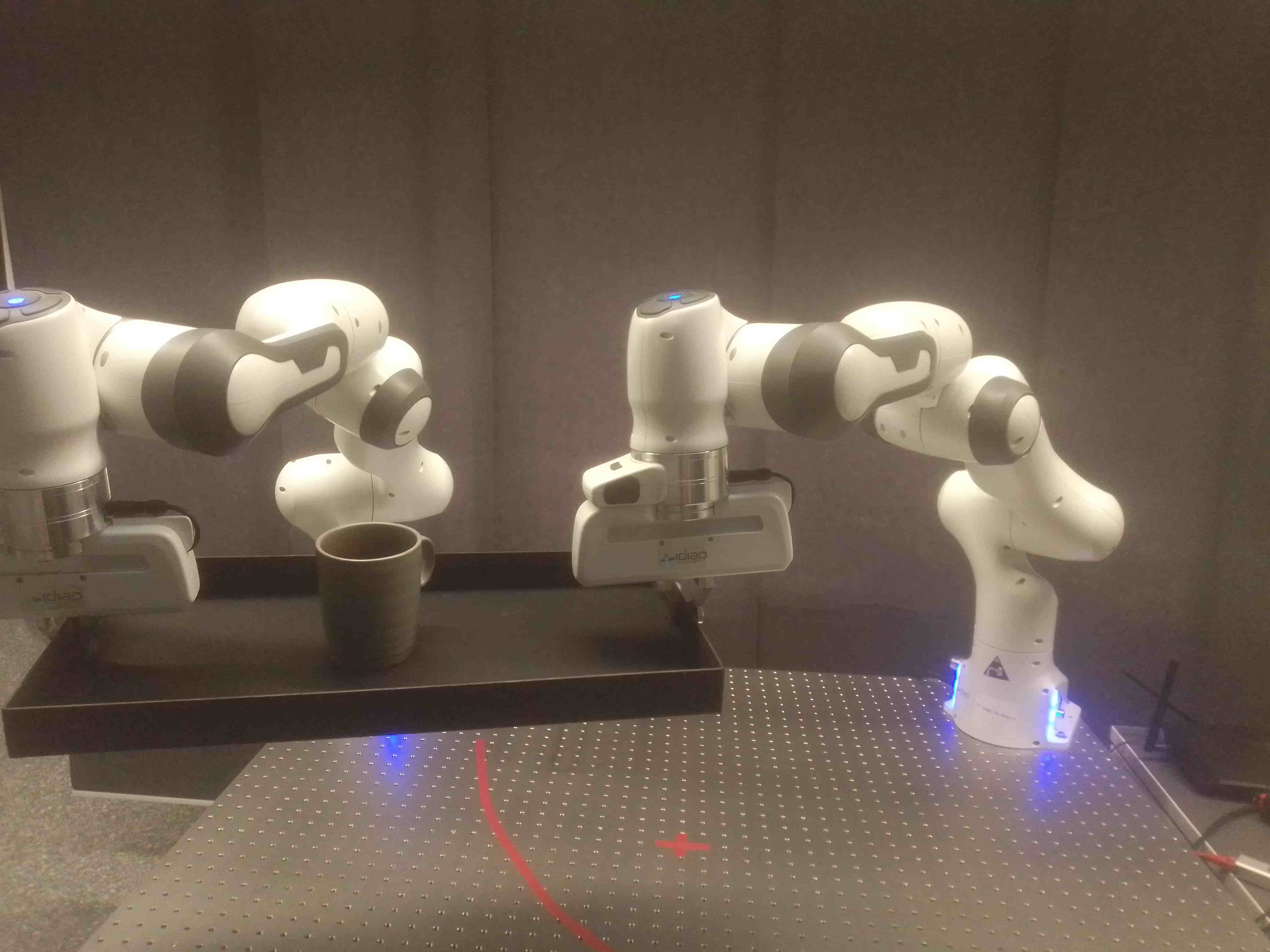}
            \end{subfigure}
            \caption{Results of balancing a plate in different configurations.}
            \label{fig:results_of_balancing_a_mug_in_different_configurations_}
        \end{figure}

        Since there are only relative constraints in this task, the robots have a compliant behavior when one of them or the plate is moved by hand. The other one then adapts its configuration accordingly. We show several different configurations in Figure \ref{fig:results_of_balancing_a_mug_in_different_configurations_}. If no external forces are applied, the robots stay in their current configuration. This experiment can also be seen in the accompanying video on our website.


\section{CONCLUSION}
\label{sec:conclusion}
    In this article we presented an extension of the cooperative dual-task space in conformal geometric algebra, namely the CGA-CDTS. This extension keeps all the benefits of the original formulation that is based on dual quaternions, but adds more tools for geometric modeling of the dual-arm tasks. After reformulating the CDTS, we showed how the cooperative pointpair can be used to simultaneously represent both end-effector positions and how that can be exploited for cooperative reaching tasks and how the additional geometric primitives facilitate the modeling of dual-arm tasks. We then demonstrated the integration of the CGA-CDTS into an existing framework for optimal control using geometric algebra. In future work the ideas of the CGA-CDTS could be used to facilitate the modeling of the task spaces of robotic hands. For a 3-finger hand, for example, the cooperative pointpair could become a cooperative circle. 

\printbibliography

@inproceedings{adornoDualPositionControl2010,
  title = {Dual Position Control Strategies Using the Cooperative Dual Task-Space Framework},
  booktitle = {2010 {{IEEE}}/{{RSJ International Conference}} on {{Intelligent Robots}} and {{Systems}}},
  author = {Adorno, Bruno Vilhena and Fraisse, Philippe and Druon, Sébastien},
  date = {2010-10},
  pages = {3955--3960},
  issn = {2153-0866},
  doi = {10.1109/IROS.2010.5650218},
  eventtitle = {2010 {{IEEE}}/{{RSJ International Conference}} on {{Intelligent Robots}} and {{Systems}}}
}

@article{adornoRobotKinematicModeling,
  title = {Robot {{Kinematic Modeling}} and {{Control Based}} on {{Dual Quaternion Algebra}} — {{Part I}}: {{Fundamentals}}.},
  author = {Adorno, Bruno Vilhena},
  pages = {48},
  langid = {english}
}

@inproceedings{almeidaCooperativeManipulationIdentification2018,
  title = {Cooperative {{Manipulation}} and {{Identification}} of a 2-{{DOF Articulated Object}} by a {{Dual-Arm Robot}}},
  booktitle = {2018 {{IEEE International Conference}} on {{Robotics}} and {{Automation}} ({{ICRA}})},
  author = {Almeida, Diogo and Karayiannidis, Yiannis},
  date = {2018-05},
  pages = {5445--5451},
  issn = {2577-087X},
  doi = {10.1109/ICRA.2018.8460511},
  eventtitle = {2018 {{IEEE International Conference}} on {{Robotics}} and {{Automation}} ({{ICRA}})}
}

@book{bayro-corrochanoGeometricAlgebraApplications2020,
  title = {Geometric {{Algebra Applications Vol}}. {{II}}: {{Robot Modelling}} and {{Control}}},
  shorttitle = {Geometric {{Algebra Applications Vol}}. {{II}}},
  author = {Bayro-Corrochano, Eduardo},
  date = {2020},
  publisher = {{Springer International Publishing}},
  location = {{Cham}},
  doi = {10.1007/978-3-030-34978-3},
  url = {http://link.springer.com/10.1007/978-3-030-34978-3},
  urldate = {2021-05-27},
  isbn = {978-3-030-34976-9 978-3-030-34978-3},
  langid = {english}
}

@article{bayro-corrochanoMotorAlgebraApproach2000,
  title = {Motor Algebra Approach for Computing the Kinematics of Robot Manipulators},
  author = {Bayro-Corrochano, Eduardo and Kähler, Detlef},
  date = {2000},
  journaltitle = {Journal of Robotic Systems},
  volume = {17},
  number = {9},
  pages = {495--516},
  issn = {1097-4563},
  doi = {10.1002/1097-4563(200009)17:9<495::AID-ROB4>3.0.CO;2-S},
  url = {https://onlinelibrary.wiley.com/doi/abs/10.1002/1097-4563%28200009%2917%3A9%3C495%3A%3AAID-ROB4%3E3.0.CO%3B2-S},
  urldate = {2021-06-15},
  langid = {english}
}

@inproceedings{defariasDesignSingularityrobustTaskpriority2017,
  title = {Design of Singularity-Robust and Task-Priority Primitive Controllers for Cooperative Manipulation Using Dual Quaternion Representation},
  booktitle = {2017 {{IEEE Conference}} on {{Control Technology}} and {{Applications}} ({{CCTA}})},
  author = {family=Farias, given=Cristiana Miranda, prefix=de, useprefix=true and Rocha, Yuri Gonçalves and Figueredo, Luis Felipe Cruz and Bernardes, Mariana Costa},
  date = {2017-08},
  pages = {740--745},
  doi = {10.1109/CCTA.2017.8062550},
  eventtitle = {2017 {{IEEE Conference}} on {{Control Technology}} and {{Applications}} ({{CCTA}})}
}

@article{fonsecaCoupledTaskSpaceAdmittance2020,
  title = {Coupled {{Task-Space Admittance Controller Using Dual Quaternion Logarithmic Mapping}}},
  author = {Fonseca, Mariana de Paula Assis and Adorno, Bruno Vilhena and Fraisse, Philippe},
  date = {2020-10},
  journaltitle = {IEEE Robotics and Automation Letters},
  volume = {5},
  number = {4},
  pages = {6057--6064},
  issn = {2377-3766},
  doi = {10.1109/LRA.2020.3010458},
  eventtitle = {{{IEEE Robotics}} and {{Automation Letters}}}
}

@article{gunnGeometricAlgebrasEuclidean2017,
  title = {Geometric {{Algebras}} for {{Euclidean Geometry}}},
  author = {Gunn, Charles},
  date = {2017-03},
  journaltitle = {Adv. Appl. Clifford Algebras},
  volume = {27},
  number = {1},
  pages = {185--208},
  issn = {0188-7009, 1661-4909},
  doi = {10.1007/s00006-016-0647-0},
  url = {http://link.springer.com/10.1007/s00006-016-0647-0},
  urldate = {2021-05-24},
  langid = {english}
}

@online{gunnProjectiveGeometricAlgebra2020,
  title = {Projective Geometric Algebra: {{A}} New Framework for Doing Euclidean Geometry},
  shorttitle = {Projective Geometric Algebra},
  author = {Gunn, Charles G.},
  date = {2020-08-18},
  eprint = {1901.05873},
  eprinttype = {arxiv},
  eprintclass = {math},
  url = {http://arxiv.org/abs/1901.05873},
  urldate = {2023-09-04},
  langid = {english},
  pubstate = {preprint}
}

@inproceedings{lahaReactiveCooperativeManipulation2021,
  title = {Reactive {{Cooperative Manipulation}} Based on {{Set Primitives}} and {{Circular Fields}}},
  booktitle = {2021 {{IEEE International Conference}} on {{Robotics}} and {{Automation}} ({{ICRA}})},
  author = {Laha, Riddhiman and Figueredo, Luis F.C. and Vrabel, Juraj and Swikir, Abdalla and Haddadin, Sami},
  date = {2021-05},
  pages = {6577--6584},
  issn = {2577-087X},
  doi = {10.1109/ICRA48506.2021.9561985},
  eventtitle = {2021 {{IEEE International Conference}} on {{Robotics}} and {{Automation}} ({{ICRA}})}
}

@inproceedings{leeSelfreconfigurableDualarmSystem1991,
  title = {A Self-Reconfigurable Dual-Arm System},
  booktitle = {1991 {{IEEE International Conference}} on {{Robotics}} and {{Automation Proceedings}}},
  author = {Lee, S. and Kim, S.},
  date = {1991-04},
  pages = {164-169 vol.1},
  doi = {10.1109/ROBOT.1991.131573},
  eventtitle = {1991 {{IEEE International Conference}} on {{Robotics}} and {{Automation Proceedings}}}
}

@article{lizaahmadshauriAssemblyManipulationSmall2011,
  title = {Assembly Manipulation of Small Objects by Dual‐arm Manipulator},
  author = {Liza Ahmad Shauri, Ruhizan and Nonami, Kenzo},
  date = {2011-01-01},
  journaltitle = {Assembly Automation},
  volume = {31},
  number = {3},
  pages = {263--274},
  publisher = {{Emerald Group Publishing Limited}},
  issn = {0144-5154},
  doi = {10.1108/01445151111150604},
  url = {https://doi.org/10.1108/01445151111150604},
  urldate = {2023-08-29}
}

@article{lowGeometricAlgebraOptimal2023,
  title = {Geometric {{Algebra}} for {{Optimal Control With Applications}} in {{Manipulation Tasks}}},
  author = {Löw, Tobias and Calinon, Sylvain},
  date = {2023},
  journaltitle = {IEEE Transactions on Robotics},
  pages = {1--15},
  issn = {1941-0468},
  doi = {10.1109/TRO.2023.3277282},
  eventtitle = {{{IEEE Transactions}} on {{Robotics}}}
}

@article{ogrenMultiObjectiveControl2012,
  title = {A {{Multi Objective Control}} Approach to {{Online Dual Arm Manipulation1}}},
  author = {Ögren, Petter and Smith, Christian and Karayiannidis, Yiannis and Kragic, Danica},
  date = {2012-01-01},
  journaltitle = {IFAC Proceedings Volumes},
  series = {10th {{IFAC Symposium}} on {{Robot Control}}},
  volume = {45},
  number = {22},
  pages = {747--752},
  issn = {1474-6670},
  doi = {10.3182/20120905-3-HR-2030.00032},
  url = {https://www.sciencedirect.com/science/article/pii/S1474667016336990},
  urldate = {2023-08-29}
}

@book{perwassGeometricAlgebraApplications2009,
  title = {Geometric Algebra with Applications in Engineering},
  author = {Perwass, Christian},
  date = {2009},
  series = {Geometry and Computing},
  number = {4},
  publisher = {{Springer}},
  location = {{Berlin}},
  isbn = {978-3-540-89067-6},
  langid = {english},
  pagetotal = {385}
}

@inproceedings{sariyildizComparisonStudyThree2011,
  title = {A Comparison Study of Three Screw Theory Based Kinematic Solution Methods for the Industrial Robot Manipulators},
  booktitle = {2011 {{IEEE International Conference}} on {{Mechatronics}} and {{Automation}}},
  author = {Sariyildiz, Emre and Temeltas, Hakan},
  date = {2011-08},
  pages = {52--57},
  publisher = {{IEEE}},
  location = {{Beijing, China}},
  doi = {10.1109/ICMA.2011.5985630},
  url = {http://ieeexplore.ieee.org/document/5985630/},
  urldate = {2023-02-22},
  eventtitle = {2011 {{IEEE International Conference}} on {{Mechatronics}} and {{Automation}} ({{ICMA}})},
  isbn = {978-1-4244-8113-2},
  langid = {english}
}

@article{smithDualArmManipulation2012,
  title = {Dual Arm Manipulation—{{A}} Survey},
  author = {Smith, Christian and Karayiannidis, Yiannis and Nalpantidis, Lazaros and Gratal, Xavi and Qi, Peng and Dimarogonas, Dimos V. and Kragic, Danica},
  date = {2012-10},
  journaltitle = {Robotics and Autonomous Systems},
  volume = {60},
  number = {10},
  pages = {1340--1353},
  issn = {09218890},
  doi = {10.1016/j.robot.2012.07.005},
  url = {https://linkinghub.elsevier.com/retrieve/pii/S092188901200108X},
  urldate = {2022-04-20},
  langid = {english}
}

@inproceedings{tassaSynthesisStabilizationComplex2012,
  title = {Synthesis and Stabilization of Complex Behaviors through Online Trajectory Optimization},
  booktitle = {2012 {{IEEE}}/{{RSJ International Conference}} on {{Intelligent Robots}} and {{Systems}}},
  author = {Tassa, Yuval and Erez, Tom and Todorov, Emanuel},
  date = {2012-10},
  pages = {4906--4913},
  issn = {2153-0866},
  doi = {10.1109/IROS.2012.6386025},
  eventtitle = {2012 {{IEEE}}/{{RSJ International Conference}} on {{Intelligent Robots}} and {{Systems}}}
}

@inproceedings{vorndammeIntegratedBiManualMotion2022a,
  title = {Integrated {{Bi-Manual Motion Generation}} and {{Control}} Shaped for {{Probabilistic Movement Primitives}}},
  booktitle = {2022 {{IEEE-RAS}} 21st {{International Conference}} on {{Humanoid Robots}} ({{Humanoids}})},
  author = {Vorndamme, Jonathan and Carvalho, João and Laha, Riddhiman and Koert, Dorothea and Figueredo, Luis and Peters, Jan and Haddadin, Sami},
  date = {2022-11},
  pages = {202--209},
  issn = {2164-0580},
  doi = {10.1109/Humanoids53995.2022.10000149},
  eventtitle = {2022 {{IEEE-RAS}} 21st {{International Conference}} on {{Humanoid Robots}} ({{Humanoids}})}
}

@inproceedings{zhuDualarmRoboticManipulation2018,
  title = {Dual-Arm Robotic Manipulation of Flexible Cables},
  booktitle = {2018 {{IEEE}}/{{RSJ International Conference}} on {{Intelligent Robots}} and {{Systems}} ({{IROS}})},
  author = {Zhu, Jihong and Navarro, Benjamin and Fraisse, Philippe and Crosnier, Andre and Cherubini, Andrea},
  date = {2018-10},
  pages = {479--484},
  publisher = {{IEEE}},
  location = {{Madrid}},
  doi = {10.1109/IROS.2018.8593780},
  url = {https://ieeexplore.ieee.org/document/8593780/},
  urldate = {2022-05-04},
  eventtitle = {2018 {{IEEE}}/{{RSJ International Conference}} on {{Intelligent Robots}} and {{Systems}} ({{IROS}})},
  isbn = {978-1-5386-8094-0},
  langid = {english}
}

\end{document}